\documentclass[letterpaper, 10 pt, conference]{ieeeconf}  
\usepackage{graphicx}
\usepackage{cite}
\usepackage{amsmath,amssymb}  
\usepackage{hyperref}  
\usepackage{url}  
\usepackage{algorithm}  
\usepackage{algpseudocode} 
\usepackage{booktabs}
\usepackage{array}
\usepackage{adjustbox}

\IEEEoverridecommandlockouts                              

\title{\LARGE \bf
DCIRNet: Depth Completion with Iterative Refinement for Dexterous Grasping of Transparent and Reflective Objects
}

\author{Guanghu Xie, Zhiduo Jiang, Yonglong Zhang,  Yang Liu\textsuperscript{\dag}, Zongwu Xie, Baoshi Cao, Hong Liu
\thanks{\textsuperscript{\dag}Corresponding author:liuyanghit@hit.edu.cn}
\thanks{*This work was supported by the Natural Science Foundation of Heilongjiang Province for Excellent Young Scholars (Grant No. YQ2024E018) and the Youth Talent Support Program of the China (Grant No. 2022-JCJQ-QT-061).}
\thanks{All authors are with with the State Key Laboratory of Robotics and Systems, Harbin Institute of Technology, Harbin 150001, Heilongjiang, China
}
}

\begin{document}
\maketitle
\thispagestyle{empty}
\pagestyle{empty}

\begin{abstract}
Transparent and reflective objects 
in everyday environments pose significant challenges for depth sensors 
due to their unique visual properties, such as specular reflections 
and light transmission. These characteristics often 
lead to incomplete or inaccurate depth estimation, 
which severely impacts downstream geometry-based vision tasks, 
including object recognition, scene reconstruction, and robotic manipulation. 
To address the issue of missing depth information in transparent 
and reflective objects, we propose DCIRNet, 
a novel multimodal depth completion network 
that effectively integrates RGB images and depth maps 
to enhance depth estimation quality.
Our approach incorporates an innovative multimodal feature fusion module 
designed to extract complementary information 
between RGB images and incomplete depth maps. 
Furthermore, we introduce a multi-stage supervision 
and depth refinement strategy that progressively improves depth completion 
and effectively mitigates the issue of blurred object boundaries.
We integrate our depth completion model into dexterous grasping frameworks 
and achieve a $44\%$ improvement in the grasp success rate for transparent and reflective objects.
We conduct extensive experiments on public datasets, 
where DCIRNet demonstrates superior performance. 
The experimental results validate the effectiveness 
of our approach and confirm its strong generalization capability 
across various transparent and reflective objects.
\end{abstract}


\begin{figure}[htbp] 
    \centering
    \includegraphics[width=\columnwidth]{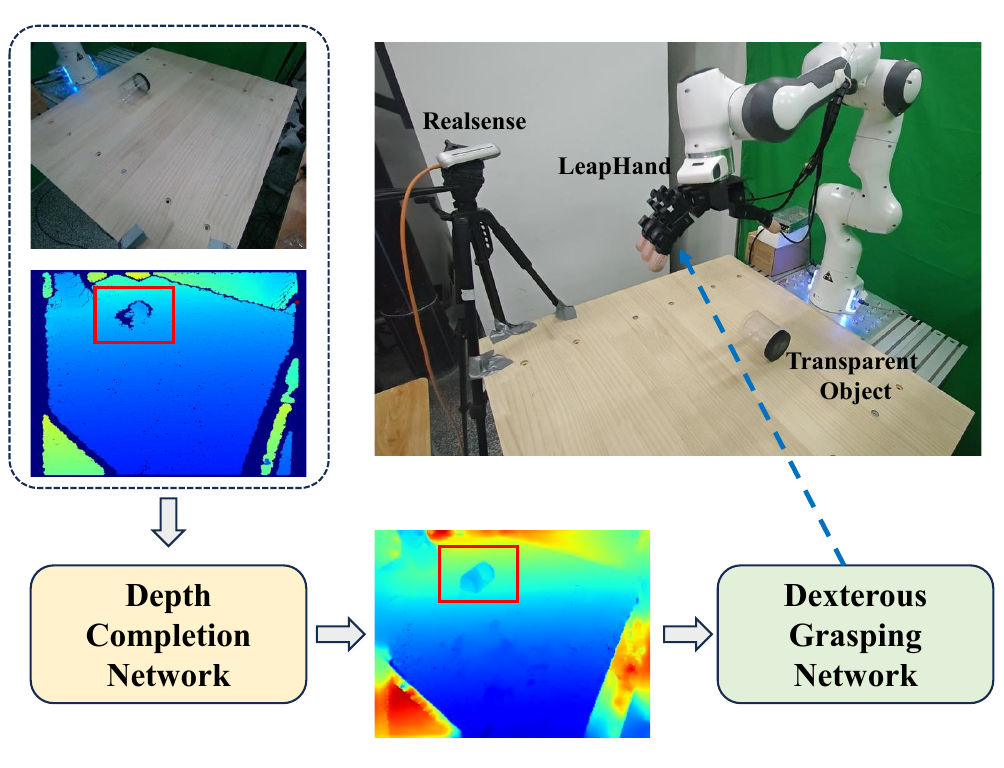} 
    \caption{The general pipeline in which a depth completion model is used to recover the depth of transparent objects, which is subsequently fed into downstream tasks.}
    \label{fig0}
\end{figure}

\section{INTRODUCTION}

Transparent and reflective objects are ubiquitous 
in our daily lives and play a crucial role in various domains, 
including industrial manufacturing, logistics, and household services. 
However, due to their inherent properties of light transmission 
and reflection, existing depth sensors struggle 
to accurately capture their depth information, 
posing significant challenges for vision-based perception 
and detection tasks\cite{cui2023light}\cite{tan2022mirror}.

Many fundamental tasks rely on complete depth information, 
and the absence of depth in transparent and reflective regions directly 
leads to incomplete input features for downstream subtasks, 
thereby compromising task execution. 
Taking dexterous grasping with multi-fingered hands as an example, 
depth completion can be used to provide more complete depth information 
for transparent and reflective objects, thereby improving the success rate of dexterous grasping, as shown in Fig.\ref{fig0}.
During dexterous grasp detection, 
the missing depth in transparent or reflective areas causes two types of detection failures. 
For fully transparent objects, the grasp detection fails entirely 
due to extensive depth loss. 
For partially transparent objects, the detector tends to focus 
only on the opaque regions while ignoring interference 
from the transparent parts, often resulting in predicted grasp poses 
that collide with the transparent regions. 
Depth completion can recover the missing depth in such areas 
and thus significantly enhance the success rate 
of downstream multi-finger dexterous grasp detection tasks.



Depth completion for transparent and reflective objects 
is a highly challenging task, as conventional depth sensing techniques 
often fail to capture accurate depth information 
due to the unique optical properties of such objects. 
Many researchers have dedicated significant efforts 
to addressing the problem of missing depth information in transparent 
and reflective surfaces to enhance the reliability 
and accuracy of vision-based perception.
Ba et al. \cite{ba2020deep} proposed a method 
that relies on specialized equipment to capture the geometric information 
of transparent and reflective surfaces. 
However, this approach lacks adaptability 
to commonly used depth sensors, such as RGB-D cameras, 
limiting its practical applicability. 
Furthermore, although multi-view methods\cite{kerr2022evo}\cite{dai2023graspnerf}
have shown promising improvements in depth estimation, 
they often introduce constraints in real-world applications, 
as they typically require multiple viewpoints. 
This makes them unsuitable for scenarios 
where only a single viewpoint is available, 
significantly reducing their feasibility 
for practical depth completion tasks.



In this work, we focus on addressing the problem of depth missing 
in transparent and reflective objects under single-view RGB-D image input. 
To this end, we propose DCIRNet, a novel multi-stage supervision 
and depth refinement model, 
which effectively fuses RGB and depth modalities to enhance depth completion. 
Our approach is designed to leverage complementary information 
between RGB images and depth maps, 
improving the robustness and accuracy of depth estimation.

The main contributions of this work are as follows:

\begin{itemize}
\item A novel multimodal feature fusion module tailored. This module facilitates effective interaction between RGB and depth modalities, enabling information complementation and significantly enhancing the network’s feature extraction capability.
\item A multi-stage supervision and depth refinement strategy, which guides the network through a coarse-to-fine depth refinement process. This hierarchical learning approach ensures progressive enhancement of depth accuracy while enforcing structural consistency.
\item Comprehensive evaluation on public datasets demonstrates that DCIRNet exhibits superior performance across multiple benchmark tests. The experimental results validate the effectiveness of our approach and confirm its strong generalization capability across various transparent and reflective objects, highlighting its practical potential for real-world depth completion tasks.
\item We applied our depth completion framework to multi-finger dexterous grasping, resulting in a $44\%$ improvement in the grasp success rate for transparent and reflective objects.
\end{itemize}

\section{RELATED WORK}
\subsection{Single-view depth completion}
Single-view depth completion has attracted significant attention 
due to its promising application potential. 
It primarily focuses on completing sparse depth maps, 
typically utilizing both RGB images 
and the corresponding sparse depth data\cite{wang2023decomposed}\cite{lin2023dyspn}. 
\cite{li2023fdct} designs a fast and accurate depth completion framework for transparent objects, 
featuring efficient fusion of low-level 
and global features through a novel architecture and loss design.
\cite{sun2024diffusion} introduces a two-stage method for depth inpainting 
of transparent and reflective objects, 
which first segments the regions and decomposes the depth loss 
into optical and geometric components, 
followed by applying diffusion-based models 
to inpaint these two types of depth separately.
\cite{fan2024tdcnet} proposes a CNN-Transformer dual-branch network 
with a multi-scale fusion module 
and a gradient-aware training strategy 
for transparent object depth completion.
\cite{dai2022domain}designs a dual-branch model based 
on Swin Transformer\cite{liu2021swin}
for RGB and depth images, 
employing a cross-attention mechanism 
for multimodal feature fusion.

\subsection{Multimodal Fusion}
Unimodal information often suffers from 
performance limitations due to its insufficient representational capacity. 
In contrast, multimodal data provide complementary 
and diverse features, which can be effectively integrated 
to enhance task performance. 
As a result, multimodal approaches 
have shown superior results in various vision tasks, 
such as autonomous driving\cite{ha2017mfnet} and semantic segmentation\cite{cao2021shapeconv}\cite{ye2019cross}. 
Multimodal feature fusion has become an active area of research, 
with many studies dedicated to designing fusion modules 
that fully leverage the complementary strengths 
of different modalities.
Cross-attention mechanisms are commonly employed 
for multimodal fusion but often incur high computational costs. 
To balance fusion performance and efficiency, recent studies\cite{jia2024geminifusion} 
introduces an innovative pixel-wise fusion module 
that leverages cross-attention 
for effective inter-modal interaction while significantly 
reducing the computational overhead.
\cite{zhang2023cmx} proposes CMX, 
an RGB-X semantic segmentation framework 
that incorporates cross-attention 
and channel-mixing modules 
to enhance global feature reasoning and alignment.

\subsection{Depth Refinement}
Depth maps obtained via direct regression 
are often affected by boundary blurring, 
leading to inaccuracies near object edges\cite{tang2024bilateral}. 
To mitigate this issue, 
depth refinement techniques are introduced, 
with most existing methods 
adopting a spatial propagation mechanism\cite{liu2017learning} 
that iteratively refines depth using local linear models.
\cite{wang2023lrru} avoids heavy feature extraction 
by first generating a coarse dense depth map 
and then iteratively refining it using spatially-variant, 
content-adaptive kernels guided by RGB and depth information.
\cite{park2020non} refines initial depth predictions 
by leveraging pixel-wise confidence 
and non-local neighbor affinities inferred 
from RGB and sparse depth inputs.
\cite{cheng2019learning} proposes CSPN, a fast and effective linear propagation model 
using recurrent convolutional operations 
with learned pixel affinities, and further enhance this approach 
in CSPN++\cite{cheng2020cspn++} 
by integrating outputs from multiple kernel sizes 
and iterative steps for improved refinement.

\section{METHOD}
The proposed depth completion model primarily 
consists of a dual-branch encoding architecture, 
a multi-modal fusion module, and a depth refinement module. 
The dual-branch encoder is designed to 
extract feature representations 
from both RGB images and depth maps. 
The multi-modal fusion module integrates information 
from both modalities to obtain fused features, 
which are subsequently fed into a decoder 
to generate initial depth predictions. 
The depth refinement module iteratively penalizes the initial predictions 
to produce more accurate and fine-grained depth maps. 
We apply supervision to both the initial depth predictions and the refined predictions, 
enabling a coarse-to-fine multi-stage supervision strategy.
Detailed descriptions of each component are provided in Sections \ref{sec3_1} to \ref{sec3_3}.

\begin{figure*}[htbp] 
    \centering
    \includegraphics[width=\textwidth]{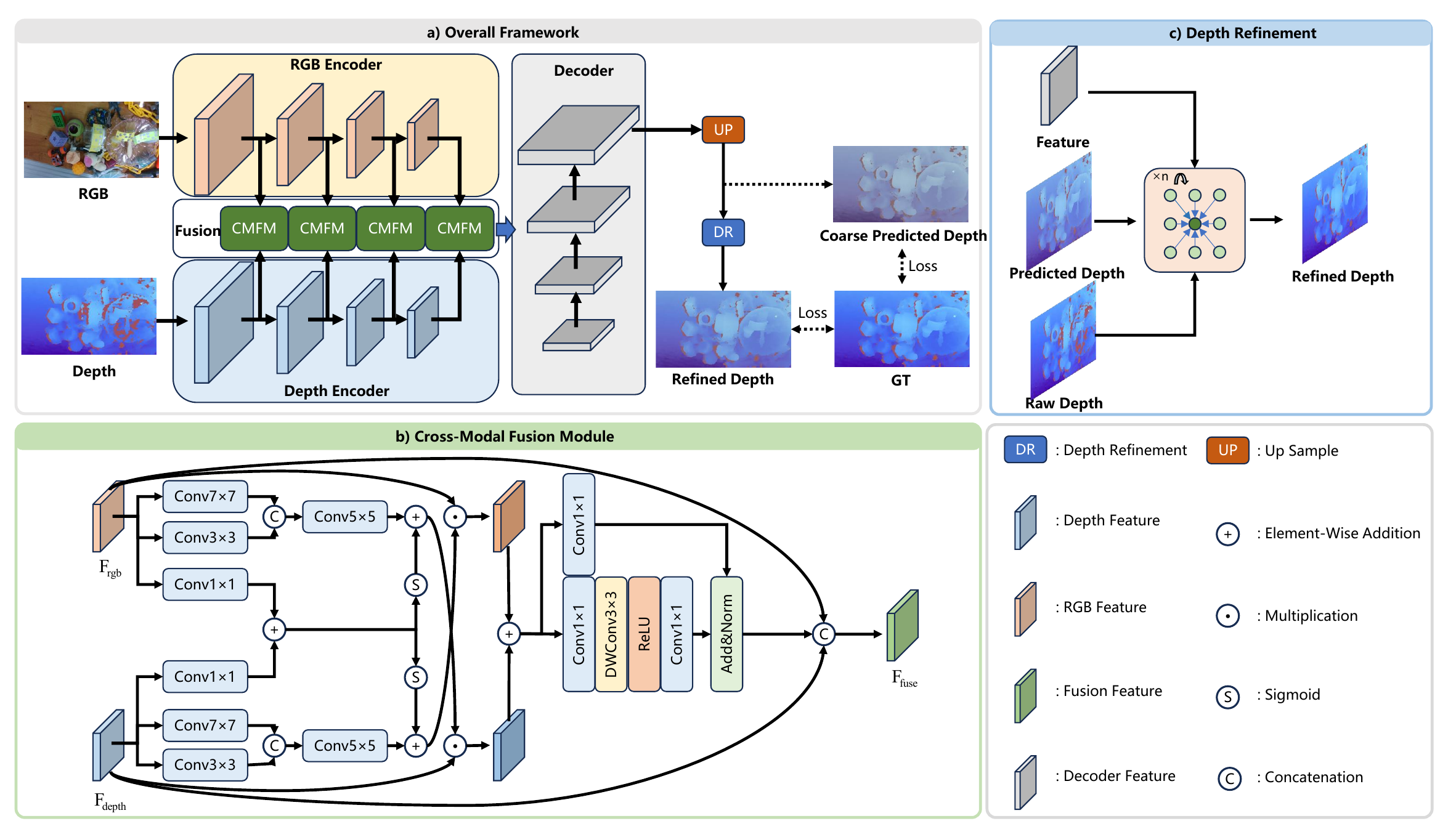} 
    \caption{DCIRNet Architecture. Our network is primarily composed of an RGB encoder, a depth encoder, a multi-modal fusion module, a decoder, and a depth refinement module. The RGB image and the depth map are first fed into their respective encoders to extract multi-stage features. These features are then fused at each stage by the proposed multi-modal fusion module. The decoder takes the fused features as input and generates an initial depth prediction. Finally, the depth refinement module performs iterative optimization based on the decoder features, the initial predicted depth, and the original sparse depth map, producing a higher-quality depth completion result.}
    \label{fig1}
\end{figure*}

\subsection{Dual-branch Architecture}
\label{sec3_1}
To extract informative features from both RGB images and depth maps, 
we adopt a dual-branch architecture in which each modality 
is independently encoded. Both branches utilize Swin Transformer 
as the backbone. Compared to conventional Transformer architectures, 
Swin Transformer reduces computational complexity 
through a shifted window mechanism 
while maintaining strong feature representation capabilities.

As illustrated in Fig. \ref{fig1}(a), the RGB and depth inputs 
are processed separately by two modality-specific Swin Transformer backbones, 
enabling the extraction of complementary features from each modality. 
These features are then fed into our proposed multi-modal fusion module 
to generate a unified representation. 
This dual-branch structure offers significantly better feature representation 
compared to single-branch designs that simply concatenate RGB 
and depth inputs before feeding them into a backbone. 
The superior performance of our approach 
is validated through additional ablation studies provided 
in subsequent sections.

After obtaining multi-scale features 
from different stages of the encoder, 
we forward them to the decoder. 
We employ UPerNet\cite{xiao2018unified} as our decoding architecture, 
which leverages a pyramid pooling module 
to capture rich global context 
and effectively integrate multi-level features 
for accurate depth prediction.

\subsection{Cross-modal Fusion Module}
\label{sec3_2}
The RGB images and raw depth maps exhibit different levels of importance 
in depth completion tasks. For instance, 
the raw depth maps typically suffer from missing depth values 
in regions containing transparent or reflective objects, 
which constitute invalid features for depth completion. 
Additionally, we argue that the significance 
of different spatial positions also varies. 
To effectively extract valuable features from each modality 
and diminish the influence of invalid features, 
we propose a novel multimodal fusion module. 
Inspired by previous studies such as \cite{zhang2023cmx} and \cite{chen2025deyolo}, 
our designed cross-modal fusion module first obtains global multimodal features 
to determine spatial importance adaptively. 
Specifically, the features from each modality 
are projected through linear layers, 
followed by pixel-wise summation to integrate the multimodal representations.

\begin{equation}
\label{equa1}
    W_{\text{fuse}} = \text{Conv}(F_{\text{rgb}}) \oplus  \text{Conv}(F_{\text{depth}})
\end{equation}

To comprehensively capture spatial features from different modalities, 
we utilize convolutional kernels of varying sizes 
to extract multi-scale features. 
These multi-scale features 
are then concatenated and projected through a linear layer.

\begin{equation}
    \label{equa2}
    \begin{aligned}
    W_{\text{rgb}} &= \text{Conv}(\text{Concat}(\text{Conv}_1(F_{\text{rgb}}),\text{Conv}_2(F_{\text{rgb}}))),\\
    W_{\text{depth}} &= \text{Conv}(\text{Concat}(\text{Conv}_1(F_{\text{depth}}),\text{Conv}_2(F_{\text{depth}})))
\end{aligned}
\end{equation}

Subsequently, we obtain the final weights 
for each modality by combining the fused features 
with their respective modality-specific features, 
as shown in Eq.\ref{equa3}:

\begin{equation}
    \label{equa3}
    \begin{aligned}
    W_{\text{rgb}} &= softmax(W_{\text{rgb}} \oplus softmax(W_{\text{fuse}})),\\
    W_{\text{depth}} &= softmax(W_{\text{depth}} \oplus softmax(W_{\text{fuse}}))
\end{aligned}
\end{equation}

Then, we multiply the weights by their corresponding modality features to obtain the enhanced features.

\begin{equation}
    \label{equa4}
    \begin{aligned}
    F_{\text{rgb}} &= F_{\text{rgb}} \odot W_{\text{depth}},\\
    F_{\text{depth}} &= F_{\text{depth}} \odot W_{\text{rgb}}
\end{aligned}
\end{equation}

Finally, we employ depth-wise convolutional layers 
to leverage the feature information from neighboring regions, 
as shown in Eq.\ref{equa5}:

\begin{equation}
    \label{equa5}
    \begin{aligned}
        F_{\text{temp}} &= \text{Conv}_{1\times 1}\left(\text{ReLU}\left(\text{DWConv}_{3\times 3}\left(\text{Conv}_{1\times 1}(F_{\text{fuse}})\right)\right)\right),\\[5pt]
        F_{\text{fuse}} &= \text{Norm}\left(F_{\text{temp}} \oplus F_{\text{fuse}}\right).
    \end{aligned}
\end{equation}

\subsection{Depth Refinement}
\label{sec3_3}
The spatial propagation module employed 
is similar to that in \cite{cheng2020cspn++}\cite{tang2024bilateral}. 
Given the original depth map \(I\), 
the features \(F_d\) output by the decoder, 
and the predicted depth map $ D^\prime $, 
we iteratively refine the predicted depth map 
according to the following equation:

\begin{equation}
    \label{equa6}
    \begin{aligned}
    D^{\prime}_{i,k,t} &= \kappa_{i,k}D^{\prime}_{i,k,t-1} + \sum_{j \in \mathcal{N}_k(i)\setminus i} \kappa_{j,k}D^{\prime}_{j,k,t-1}, \\[8pt]
    \kappa_{i,k} &= 1 - \sum_{j \in \mathcal{N}_k(i)\setminus i}\kappa_{j,k}, \\[8pt]
    \kappa_{j,k} &= \frac{\kappa^{\prime}_{j,k}}{\sum\limits_{j \in \mathcal{N}_k(i)\setminus i}|\kappa^{\prime}_{j,k}|},
    \end{aligned}
\end{equation}

In the equation above, $\kappa$ represents 
an affinity map determined by image content, 
while $\kappa^{\prime}$ is produced 
by convolutional layers operating on the decoder features $F_d$. The indices i and j denote the i-th pixel 
and its corresponding j-th neighboring pixel, respectively.
Additionally, enforcing an $l^1$-norm constraint on $\kappa^{\prime}$ ensures numerical stability 
during the iterative propagation \cite{cheng2019learning}. 
$\mathcal{N}_k$ specifically 
denotes the adjacent pixels within a $k \times k$ local window, 
independent of the original depth measurement validation.

Then, the original depth map 
is embedded into the spatial propagation mechanism 
for iterative refinement, as shown in the following equation:

\begin{equation}
    D^{\prime}_{i,k,t} = (1 - \varphi_{i,k}\mathbb{I}(I_i))D^{\prime}_{i,k,t} + \varphi_{i,k}\mathbb{I}(I_i),
\end{equation}

Here, $\varphi_{i,k}$ denotes the confidence value, and $\mathbb{I}$ is used to extract valid values from the depth map.

The weights vary across different kernels and iteration steps, 
and the overall depth value after iteration 
is obtained by the following equation:

\begin{equation}
    D^{\prime}_i = \sum_{t \in \mathcal{T}} \sum_{k \in \mathcal{K}} \alpha _{i,t} \beta_{i,k} D^{\prime}_{i,k,t}.
\end{equation}

Here, $\alpha_{i,t}$ and $\beta_{i,k}$ represent the weights corresponding 
to different iteration steps and kernel sizes, respectively. 
The set $\mathcal{K}$ refers to kernel sizes, 
typically selected from $\{3, 5, 7\}$ 
to represent varying receptive fields. 
The set $\mathcal{T}$ denotes temporal steps 
within the propagation process, commonly chosen as $\{0, \lfloor T/2 \rfloor, T\}$ 
to reflect multi-stage iterative refinement.

It is worth noting that we apply supervision to both the coarse depth predictions and the iteratively refined depth values, thereby implementing a two-stage supervision scheme that progresses from coarse to fine.

\subsection{Loss Function}
\label{sec3_4}
Both the coarse prediction $\tilde{\mathcal{D}}$ and the refined output $\hat{\mathcal{D}}$ of the model are subjected to supervision, as defined by the following equation:

\begin{equation}
    \mathcal{L} = \omega_{\tilde{\mathcal{D}}} \mathcal{L}_{\tilde{\mathcal{D}}} + \omega_{\hat{\mathcal{D}}} \mathcal{L}_{\hat{\mathcal{D}}},
\end{equation}

where $\mathcal{L}$ is the total loss, $\mathcal{L}_{\tilde{\mathcal{D}}}$ and $\mathcal{L}_{\hat{\mathcal{D}}}$ denote the losses from the coarse and refined depth predictions respectively, and $\omega_{\tilde{\mathcal{D}}}$ and $\omega_{\hat{\mathcal{D}}}$ are their corresponding weights.

Both the coarse prediction loss and the refined loss consist of three components: depth loss, normal loss, and gradient loss, and can be formulated as:

\begin{equation}
    \mathcal{L}_i = \omega_n \mathcal{L}_n + \omega_d \mathcal{L}_d + \omega_g \mathcal{L}_g,
\end{equation}

where $\mathcal{L}_i$ denotes the total loss computed on depth map $i$, which can be either the initial prediction $\tilde{D}$ or the refined prediction $\hat{D}$. It consists of the normal loss $\mathcal{L}_n$, the depth loss $\mathcal{L}_d$, and the gradient loss $\mathcal{L}_g$, weighted by the corresponding coefficients $\omega_n$, $\omega_d$, and $\omega_g$, respectively.


\section{EXPERIMENT}
\subsection{Datasets}
We evaluate our method on the DREDS\cite{dai2022domain} and TransCG datasets\cite{fang2022transcg}.
The DREDS dataset includes two subsets: DREDS-CatKnown, containing over 100k RGB-D images of 1,801 objects from 7 categories 
with diverse materials, 
and DREDS-CatNovel, with 11.5k images of 
60 novel-category objects to evaluate cross-category generalization 
under challenging materials.
The TransCG dataset consists of 57,715 RGB-D images 
from 130 scenes with diverse backgrounds, captured using two cameras, 
and is divided into 34,191 training and 23,524 testing samples.

\subsection{Metrics}
We evaluate the proposed depth completion model using four commonly adopted metrics, including RMSE, REL, MAE, and threshold accuracy $\delta$. These metrics are defined as follows:

\begin{itemize}
    \item \textbf{RMSE}: Root mean squared error between predicted and ground-truth depth.
    \item \textbf{REL}: Mean absolute relative error between predicted and ground-truth depth.
    \item \textbf{MAE}: Mean absolute error between predicted and ground-truth depth.
    \item \textbf{Threshold Accuracy $\delta$}: Percentage of pixels satisfying 
    $\max\left(\frac{d}{d^*}, \frac{d^*}{d}\right) < \delta$
    ,where $d$ and $d^*$ denote the predicted and ground-truth depth, respectively. The thresholds $\delta$ used in our experiments are set to 1.05, 1.10, and 1.25.
\end{itemize}

\subsection{Implementation Details}
The hardware used in our experiments includes Intel Xeon 8358P CPU and Nvidia RTX 4090 GPU.
We train our model using the AdamW optimizer 
with an initial learning rate of \(0.0001\), 
for \(20\) epochs, and a batch size of \(4\). 
Input images are resized to \(224 \times 224\) pixels 
before being fed into the model. 
For evaluation, we adhere to dataset-specific configurations. 
For example, images from the DREDS dataset 
are resized to \(224 \times 126\), 
while those from the TransCG dataset are resized to \(240 \times 320\).

\subsection{Experimental Results}
\subsubsection{DREDS Datasets}
Following the experimental protocol established in \cite{dai2022domain}, 
we trained our proposed model on the training set of the DREDS-CatKnown dataset 
and conducted comprehensive evaluations on both the DREDS-CatKnown test set 
and the DREDS-CatNovel dataset. As quantitatively demonstrated in Tab.\ref{tab1}, 
our method achieves superior performance compared to NLSPN and LIDF baselines 
on the DREDS-CatKnown test set, while attaining comparable results 
with the reference approach \cite{dai2022domain}. More notably, the proposed method exhibits 
enhanced generalization capability by outperforming all compared methods, 
including \cite{dai2022domain}, on the more challenging DREDS-CatNovel test set 
that contains novel object categories.
To qualitatively validate our findings, 
we provide visual comparisons of prediction results 
from both DREDS-CatKnown and DREDS-CatNovel test sets. 
These visualization results effectively 
demonstrate the superior generalization capability 
and robustness of our approach across different testing scenarios. 
Although the SwinDRNet method achieves comparable performance to ours 
on the DREDS-CatKnown dataset in terms of quantitative metrics, 
visual comparisons reveal that our approach 
more effectively addresses the issue of blurred edges 
and contours in depth completion. 
This improvement is particularly significant for downstream tasks, 
such as perceiving target objects in cluttered environments, 
where minimizing background interference is crucial.

\begin{table}[htbp]
    \centering
    \caption{Performance comparison on DREDS dataset.}
    \label{tab1}
    \resizebox{\columnwidth}{!}{
    \begin{tabular}{lccc ccc}
    \toprule
    Methods & RMSE$\downarrow$ & REL$\downarrow$ & MAE$\downarrow$ & $\delta_{1.05}\uparrow$ & $\delta_{1.10}\uparrow$ & $\delta_{1.25}\uparrow$ \\
    \midrule
    \multicolumn{7}{c}{DREDS-CatKnown}\\
    \midrule
    NLSPN\cite{park2020non} & \textbf{0.010} & 0.009 & 0.006 & 97.48 & 99.51 & 99.97 \\
    LIDF\cite{zhu2021rgb} & 0.016 & 0.018 & 0.011 & 93.60 & 98.71 & 99.92 \\
    SwinDRNet\cite{dai2022domain} & $\mathbf{0.010}$ & 0.008 & $\mathbf{0.005}$ & $\mathbf{98.04}$ & $\mathbf{99.62}$ & $\mathbf{99.98}$ \\
    DCIRNet(ours) & 0.011 & $\mathbf{0.007}$ & $\mathbf{0.005}$ & 97.65 & 99.30 & 99.95 \\
    \midrule
    \multicolumn{7}{c}{DREDS-CatNovel}\\
    \midrule
    NLSPN\cite{park2020non} & 0.026 & 0.039 & 0.015 & 78.90 & 89.02 & 97.86 \\
    LIDF\cite{zhu2021rgb} & 0.082 & 0.183 & 0.069 & 23.70 & 42.77 & 75.44 \\
    SwinDRNet\cite{dai2022domain} & 0.022 & 0.034 & 0.013 & 81.90 & 92.18 & 98.39 \\
    DCIRNet(ours) & $\mathbf{0.021}$ & $\mathbf{0.031}$ & $\mathbf{0.012}$ & $\mathbf{83.37}$ & $\mathbf{92.66}$ & $\mathbf{98.43}$ \\
    \bottomrule
    \end{tabular}}
\end{table}

\begin{figure}[htbp] 
    \centering
    \includegraphics[width=\columnwidth]{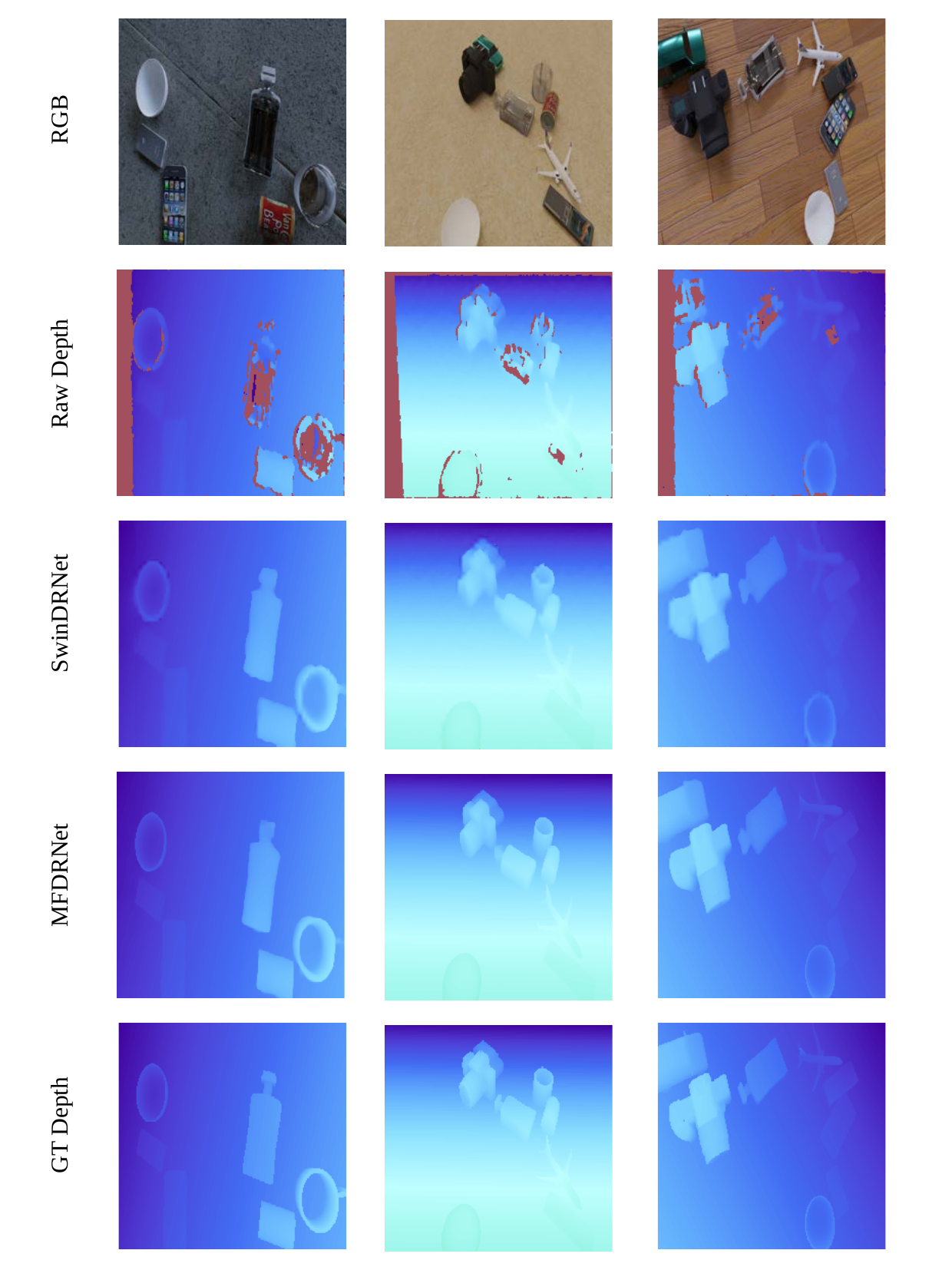} 
    \caption{Depth Completion Visualizations of Different Models on the DREDS-CatKnown Dataset}
    \label{fig2}
\end{figure}

\begin{figure}[htbp] 
    \centering
    \includegraphics[width=\columnwidth]{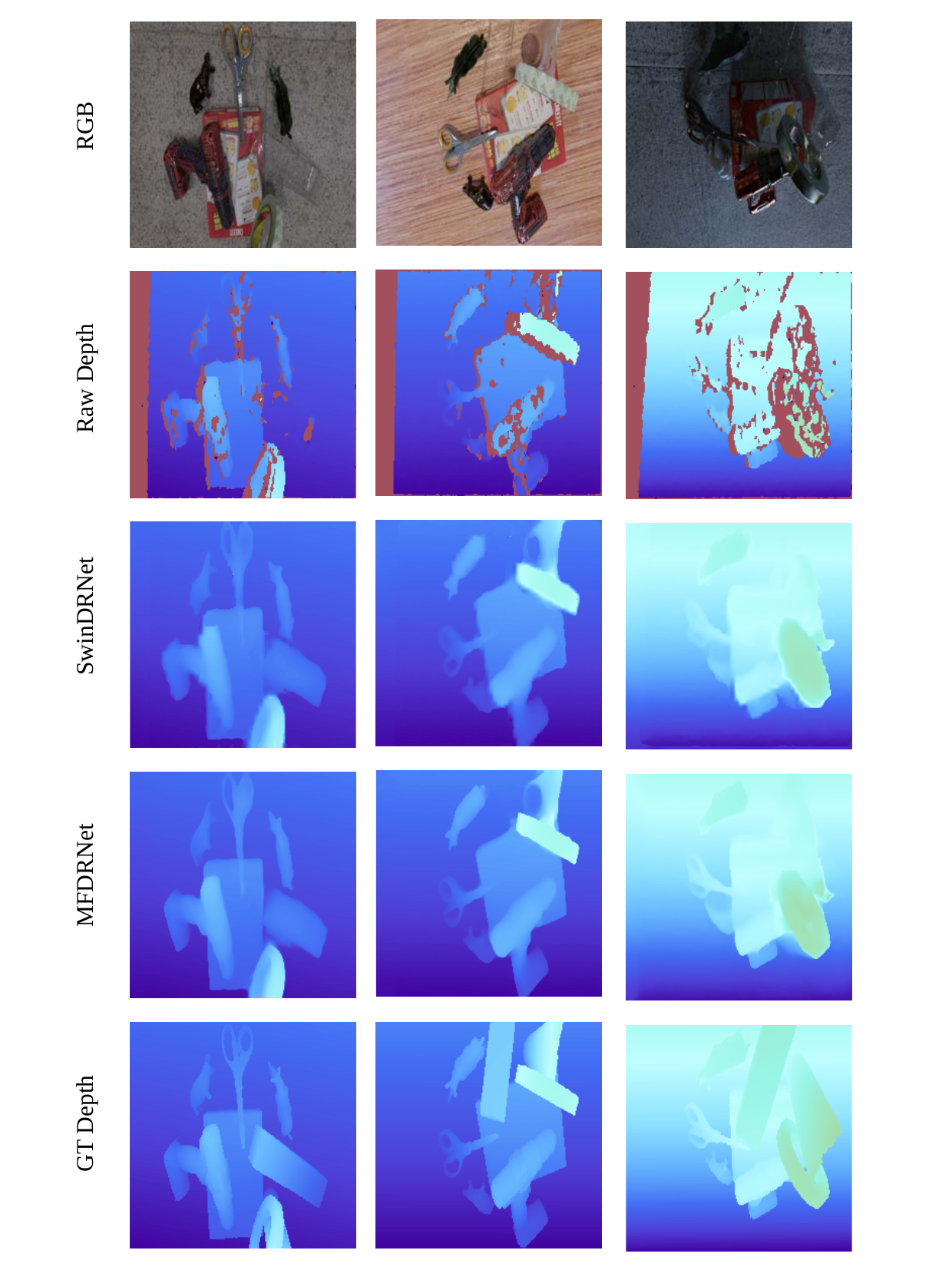} 
    \caption{Depth Completion Visualizations of Different Models on the DREDS-CatNovel Dataset}
    \label{fig3}
\end{figure}

\subsubsection{TransCG Datasets}
\begin{figure}[htbp] 
    \centering
    \includegraphics[width=\columnwidth]{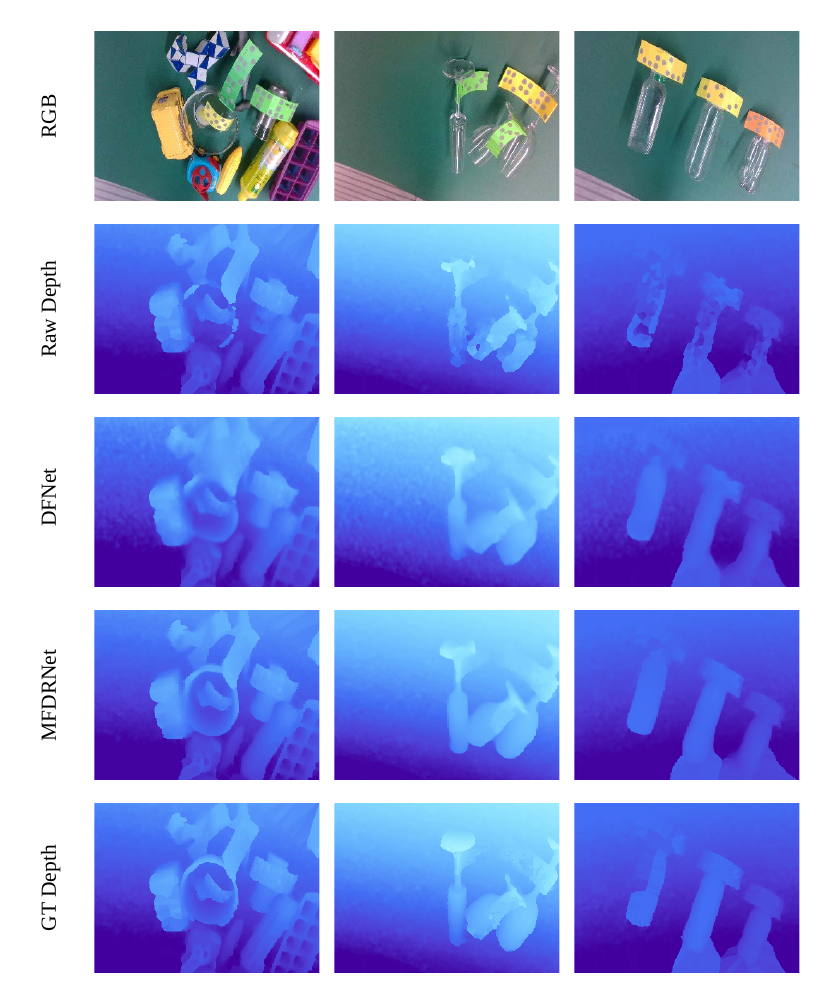} 
    \caption{Depth Completion Visualizations of Different Models on the TransCG Dataset}
    \label{fig4}
\end{figure}

We train our model on the training split of the TransCG dataset 
and conduct systematic performance evaluation on its official test set.
Following the setting in~\cite{fang2022transcg}, we constrain the valid depth range to $[0.3, 1.5]$ during the loss computation. 
The experimental results (detailed in Tab.\ref{tab2}) 
demonstrate that our method significantly 
surpasses numerous existing approaches.
Furthermore, 
We visualize the depth completion results of our model, 
as illustrated in Fig.\ref{fig4}. The figure demonstrates the model’s effectiveness 
in addressing the issue of blurred object boundaries. 
Our approach exhibits strong generalization capabilities, 
achieving satisfactory completion performance even in the absence 
of ground truth depth labels, 
as evidenced by the rightmost set of images in Fig.\ref{fig4}.

\begin{table}[htbp]
    \centering
    \caption{Performance comparison of different methods on TransCG dataset}
    \label{tab2}
    \resizebox{\columnwidth}{!}{
    \begin{tabular}{lcccccc}
    \toprule
    \textbf{Methods} & \textbf{RMSE} $\downarrow$ & \textbf{REL} $\downarrow$ & \textbf{MAE} $\downarrow$ & \textbf{$\delta_{1.05}$}$\uparrow$ & \textbf{$\delta_{1.10}$}$\uparrow$ & \textbf{$\delta_{1.25}$}$\uparrow$ \\
    \midrule
    CG\cite{sajjan2020clear}         & 0.054  & 0.083  & 0.037  & 50.48  & 68.68  & 95.28  \\
    DFNet\cite{fang2022transcg}      & 0.018  & 0.027  & 0.012  & 83.76  & 95.67  & 99.71  \\
    LIDF\cite{zhu2021rgb} & 0.019  & 0.034  & 0.015  & 78.22  & 94.26  & 99.80  \\
    TCRNet\cite{zhai2024tcrnet}     & 0.017  & 0.020  & 0.010  & 88.96  & 96.94  & $\mathbf{99.87}$  \\
    TranspareNet\cite{xu2021seeing} & 0.026  & 0.023  & 0.013  & 88.45  & 96.25  & 99.42  \\
    FDCT\cite{li2023fdct}       & 0.015  & 0.022  & 0.010  & 88.18  & 97.15  & 99.81  \\
    TODE-Trans\cite{chen2023tode}  & $\mathbf{0.013}$  & 0.019  & $\mathbf{0.008}$  & 90.43  & 97.39  & 99.81  \\
    \midrule
    DCIRNet (ours) & 0.015 & $\mathbf{0.018}$ & $0.009$ & $\mathbf{91.53}$ & $\mathbf{97.49}$ & $99.86$ \\
    \bottomrule
    \end{tabular}}
\end{table}

\subsubsection{Dexterous grasping experiment}
\begin{figure}[htbp] 
    \centering
    \includegraphics[width=\columnwidth]{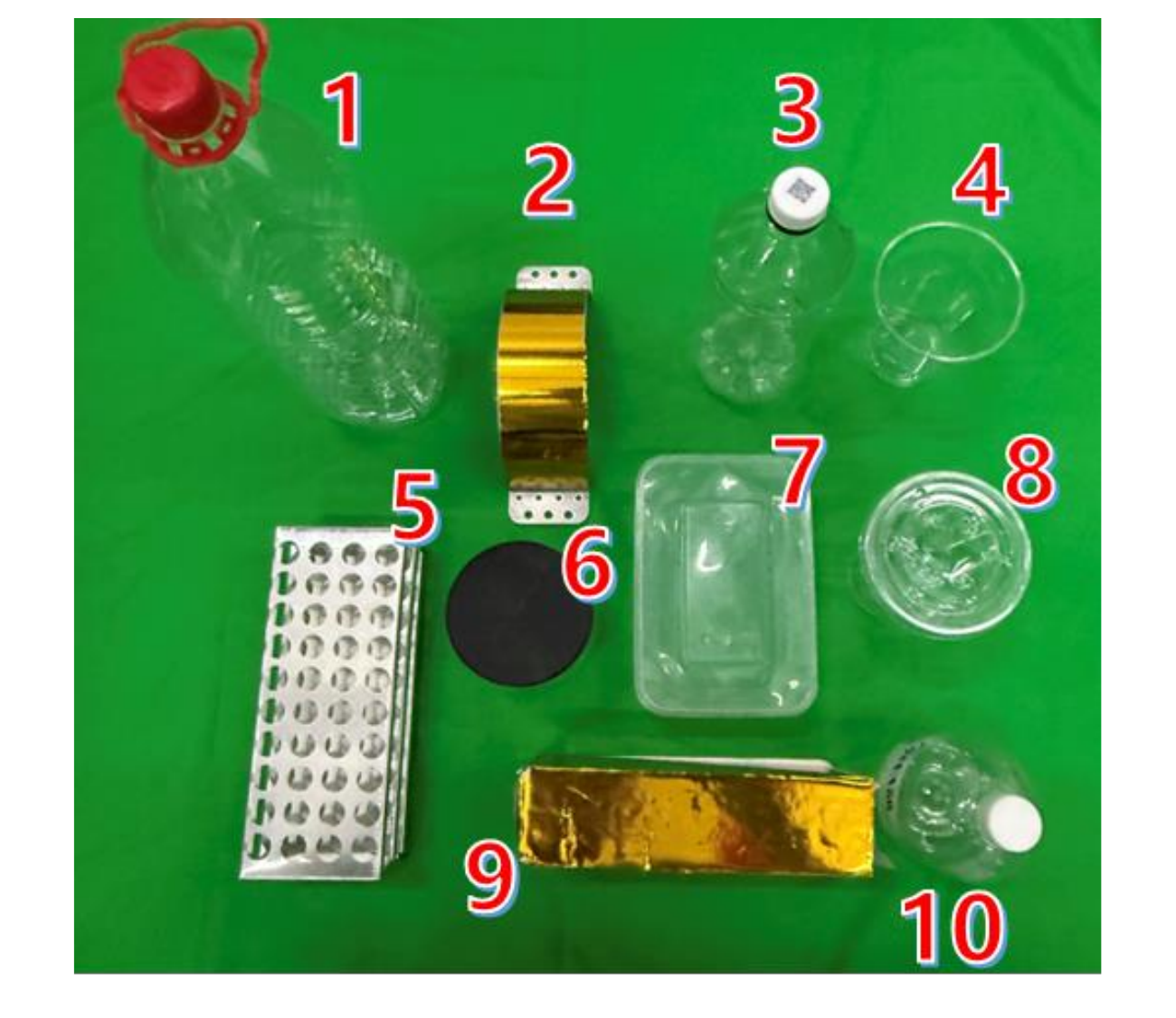} 
    \caption{Objects in the real-world grasp experiment.1.Mineral Water Bottle;2.Metal Component;3.Carbonated Drink Bottle;
    4.Drinking Cup;5.Test Tube Rack;6.Storage Plastic Bottle;7.Lunch Box;8.Lidded Coffee Cup;9.Reflective Foam;10.Hand Sanitizer Bottle.}
    \label{fig_objs}
\end{figure}

We integrate our proposed depth completion method 
into the front-end of the multi-finger dexterous grasping framework DexGraspNetV2\cite{zhang2024dexgraspnet} 
and conduct grasping experiments on transparent and reflective objects. The target objects are shown in Fig.\ref{fig_objs}, and the experimental results are summarized in Tab.\ref{table_grasp}. The results demonstrate that incorporating depth completion significantly improves the grasp success rate of DexGraspNetV2 when handling transparent and reflective objects, thereby highlighting the practical value of the proposed depth completion approach.

\begin{figure}[htbp] 
    \centering
    \includegraphics[width=\columnwidth]{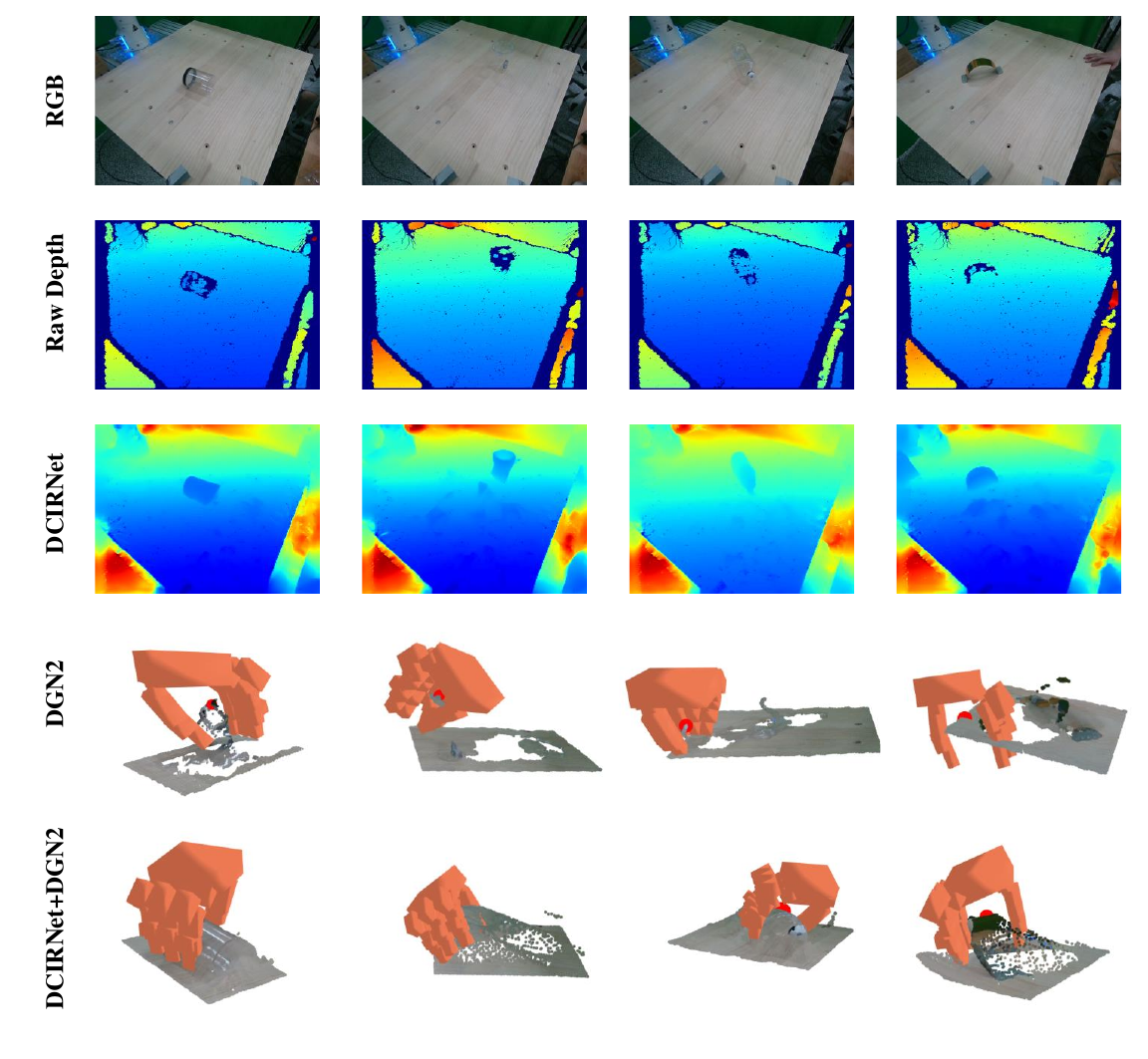} 
    \caption{Depth Completion and DexGrasp Visualizations}
    \label{fig5}
\end{figure}

\begin{figure}[htbp] 
    \centering
    \includegraphics[width=\columnwidth]{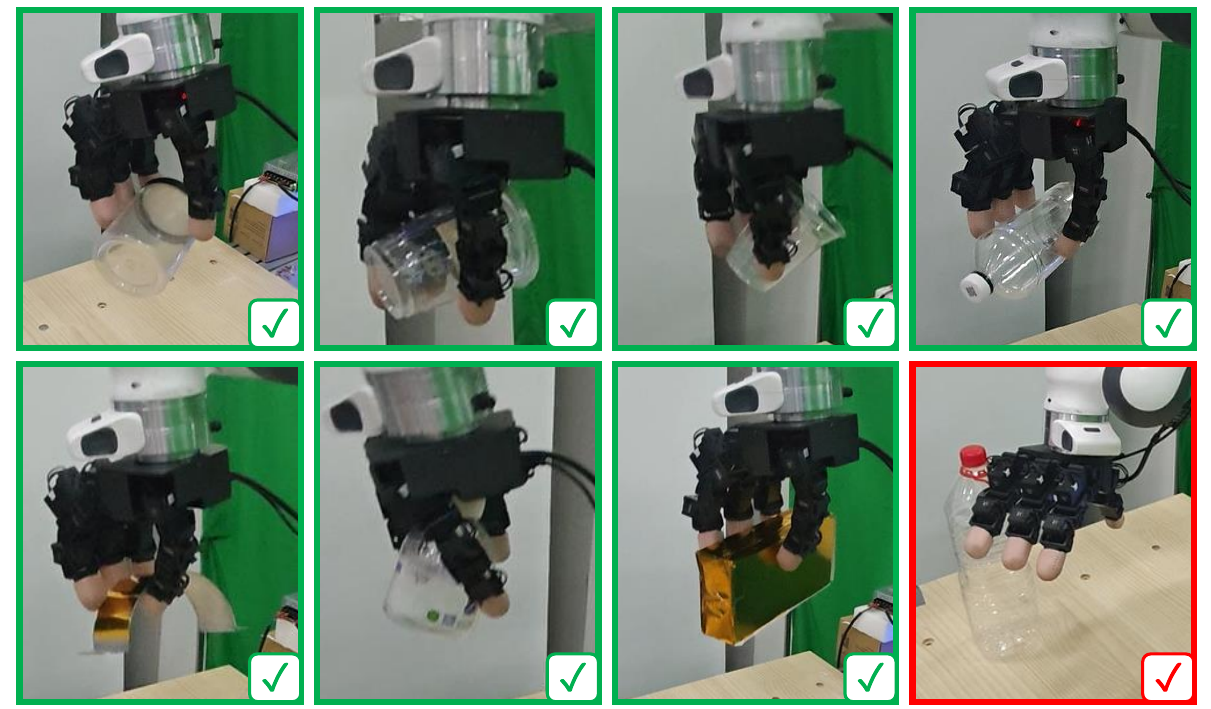} 
    \caption{Real-world dexterous grasping examples. Green indicates successful grasps, and red indicates failed grasps.}
    \label{fig6}
\end{figure}

\begin{table}[ht]
    \centering
    \caption{Grasping experiments in real-world scenes}
    \resizebox{\columnwidth}{!}{
    \begin{tabular}{lcccc}
    \toprule
    
\textbf{Objs} & \textbf{DexGraspNetV2} & \textbf{DCIRNet+DexGraspNetV2} \\
\midrule
1. Mineral Water Bottle     & 2/5 & 4/5 \\
2. Metal Component          & 3/5 & 3/5 \\
3. Carbonated Drink Bottle  & 1/5 & 5/5 \\
4. Drinking Cup             & 0/5 & 3/5 \\
5. Test Tube Rack           & 1/5 & 4/5 \\
6. Storage Plastic Bottle   & 1/5 & 5/5 \\
7. Lunch Box                & 2/5 & 4/5 \\
8. Lidded Coffee Cup        & 2/5 & 4/5 \\
9. Reflective Foam          & 3/5 & 5/5 \\
10. Hand Sanitizer Bottle   & 4/5 & 4/5 \\
\midrule
\textbf{Success Rate}       & $38.00\%$ & \textbf{82.00\%} \\ \bottomrule

    \end{tabular}
    }
    \label{table_grasp}
    \end{table}

\subsection{Ablation Studies}
We conducted additional experiments 
to further investigate the effects of the cross modal fusion modules(CMFM) and the depth refinement. 
The detailed results are described as follows:
\subsubsection{Effectiveness of the dual-branch structure}
We first combine the RGB and depth maps and input them into a single-branch backbone, 
which is a commonly used structure in previous depth completion works\cite{fang2022transcg}\cite{li2023fdct}, 
as a baseline. This is used to demonstrate the effectiveness 
of the dual-branch structure with the multimodal fusion module 
that we design.
As shown in Tab.\ref{tab3}, 
integrating multimodal information 
with our designed fusion module significantly 
enhanced the model's performance. 
This indicates that our multimodal fusion module 
effectively captures essential complementary information 
from both RGB images and depth maps, playing a crucial role 
in improving depth completion performance.


\begin{table}[htbp]
    \centering
    \caption{Performance comparison of different methods on DREDS-CatNovel dataset}
    \label{tab3}
    \resizebox{\columnwidth}{!}{
    \begin{tabular}{ccccccccc}
    \toprule
    \textbf{CMFM} & \textbf{DR} & \textbf{RMSE} $\downarrow$ & \textbf{REL} $\downarrow$ & \textbf{MAE} $\downarrow$ & \textbf{$\delta_{1.05}$}$\uparrow$ & \textbf{$\delta_{1.10}$}$\uparrow$ & \textbf{$\delta_{1.25}$}$\uparrow$ \\
    \midrule
    &  & 0.022  & 0.039  & 0.015  & 79.57  & 92.31  & 98.48  \\
    $\surd$ &  & $0.022$  & 0.037  & $0.014$  & 80.95  & $92.34$  & $98.32$  \\
    $\surd$ &$\surd$  & $\mathbf{0.021}$ & $\mathbf{0.031}$ & $\mathbf{0.012}$ & $\mathbf{83.37}$ & $\mathbf{92.66}$ & $\mathbf{98.43}$  \\
    \bottomrule
    \end{tabular}}
\end{table}

\section{CONCLUSIONS}
In this work, we have proposed 
a dual-branch multi-stage refinement supervision network tailored 
for depth completion of transparent and reflective objects. 
The proposed model has been extensively evaluated 
on publicly available datasets, and experimental results 
have demonstrated the significant effectiveness 
of our multimodal fusion module 
and multi-stage depth refinement supervision strategy. 
Our method effectively addresses the issue 
of blurred object boundaries in the depth completion task 
for transparent and reflective objects.
Our method achieved superior performance compared 
to numerous existing approaches, 
indicating robust generalization capability and effectiveness.
Additionally, our method is effectively applied to the dexterous grasping of transparent and reflective objects, increasing the success rate of grasping such objects by $44\%$. 
In future studies, we aim to further optimize the network 
towards a lightweight design, 
striving to achieve an optimal balance between accuracy 
and computational efficiency.

\bibliographystyle{IEEEtran}
\bibliography{IEEEabrv, refs}

\end{document}